\pdfoutput=1

\documentclass[10pt, conference, compsocconf]{IEEEtran}
\usepackage{placeins}
\usepackage[normalem]{ulem}

\usepackage{listings}
\usepackage{xcolor}
\definecolor{codegreen}{rgb}{0,0.6,0}
\definecolor{codegray}{rgb}{0.5,0.5,0.5}
\definecolor{codepurple}{rgb}{0.58,0,0.82}
\definecolor{backcolour}{rgb}{0.95,0.95,0.92}

\lstdefinestyle{mystyle}{
    backgroundcolor=\color{backcolour},   
    commentstyle=\color{codegreen},
    keywordstyle=\color{magenta},
    numberstyle=\tiny\color{codegray},
    stringstyle=\color{codepurple},
    basicstyle=\ttfamily\footnotesize,
    breakatwhitespace=false,         
    breaklines=true,                 
    captionpos=b,                    
    keepspaces=true,                 
    numbers=left,                    
    numbersep=5pt,                  
    showspaces=false,                
    showstringspaces=false,
    showtabs=false,                  
    tabsize=2
}
\usepackage{cite}
\usepackage{graphicx}

\ifCLASSINFOpdf
\else
\fi
\usepackage[cmex10]{amsmath}

\usepackage{stfloats}
\usepackage{url}
\begin{document}
\title{Generate to Understand for Representation}


% \author{\IEEEauthorblockN{Authors Name/s per 1st Affiliation (Author)}
% \IEEEauthorblockA{line 1 (of Affiliation): dept. name of organization\\
% line 2: name of organization, acronyms acceptable\\
% line 3: City, Country\\
% line 4: Email: name@xyz.com}
% \and
% \IEEEauthorblockN{Authors Name/s per 2nd Affiliation (Author)}
% \IEEEauthorblockA{line 1 (of Affiliation): dept. name of organization\\
% line 2: name of organization, acronyms acceptable\\
% line 3: City, Country\\
% line 4: Email: name@xyz.com}
% }

% \author{\IEEEauthorblockN{Changshang Xue \footnote{Work at Baidu.Inc}}
%   \IEEEauthorblockA{ xuechangshang@baidu.com* }
% }

\author{\IEEEauthorblockN{1st Changshang Xue*  }
  \IEEEauthorblockA{
    Baidu Inc.\\
    laohur@gmail.com
  }
  \and
  \IEEEauthorblockN{2nd Xiande Zhong}
  \IEEEauthorblockA{
    Baidu Inc.\\
    zhongxiande@baidu.com
  }
  \and
  \IEEEauthorblockN{3rd Xiaoqing Liu}
  \IEEEauthorblockA{
    Baidu Inc.\\
    liuxiaoqing02@baidu.com
  }
}


%Telephone: (800) 555--1212, Fax: (888) 555--1212}


\maketitle

\begin{abstract}
  %auto-ignore
\label{sec:abstract}
In recent years, a significant number of high-quality pretrained models have emerged, greatly impacting Natural Language Understanding (NLU), Natural Language Generation (NLG), and Text Representation tasks. Traditionally, these models are pretrained on custom domain corpora and finetuned for specific tasks, resulting in high costs related to GPU usage and labor. Unfortunately, recent trends in language modeling have shifted towards enhancing performance through scaling, further exacerbating the associated costs.

Introducing GUR: a pretraining framework that combines language modeling and contrastive learning objectives in a single training step. We select similar text pairs based on their Longest Common Substring (LCS) from raw unlabeled documents and train the model using masked language modeling and unsupervised contrastive learning. The resulting model, GUR, achieves impressive results without any labeled training data, outperforming all other pretrained baselines as a retriever at the recall benchmark in a zero-shot setting. Additionally, GUR maintains its language modeling ability, as demonstrated in our ablation experiment. Our code is available at \url{https://github.com/laohur/GUR}.

\end{abstract}

\begin{IEEEkeywords}
  self-supervised pre-train; contrastive learning; language model; zero-shot learning; text representation; NLP; NLU; NLG; retrieval
\end{IEEEkeywords}

% \fi
\IEEEpeerreviewmaketitle

%auto-ignore
\section{Introduction}
\label{sec:intro}
\begin{figure*}[h]
    \centering
    \includegraphics[width=\textwidth]{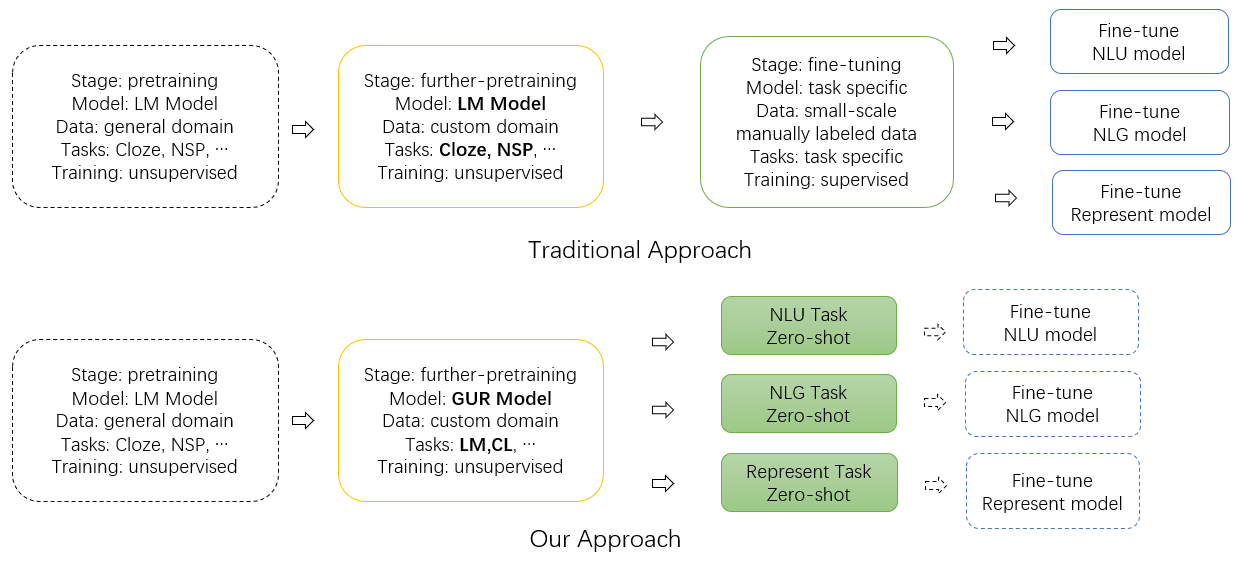}
    \caption{Comparison between the traditional pretraining-finetuning paradigm and our proposed framework GUR: Instead of domain-adaptive further-pretraining using only the LM objective and finetuning on various downstream tasks, we post-pretrain the model with both LM and CL objectives on a custom domain and employ zero-shot learning for NLU, NLG, and recall tasks. The model "GUR-CL" follows the traditional approach without the CL objective. The other models utilize the second approach. Without additional resources, the "GUR-FULL" model maintains the same LM capabilities as the traditional approach model while acquiring the ability for text representation using our approach. All models are initialized from a pretrained LM on a general corpus, minimizing costs.}
\end{figure*}

Pre-training methods that learn directly from raw text have revolutionized NLP and related fields in recent years. Neural networks from the transformer \cite{transformer} family are trained on large general corpora for self-supervised, task-agnostic objectives, such as autoregressive and masked language modeling. Subsequently, these networks are fine-tuned on a small amount of labeled data for various downstream tasks. This pretraining-finetuning pipeline has significantly enhanced the performance of numerous NLP tasks, including NLU, NLG, and text representation. However, the specific training datasets required for these tasks can be expensive or insufficient. Consequently, the costs of GPUs and labor soar for scaling these models.

Despite numerous efforts to study and improve the efficiency of language model pretraining, the majority of these efforts concentrate on specific aspects of the traditional framework. The development of "text-to-text" \cite{T5} as a standardized input-output interface has facilitated task-agnostic architectures for NLU and NLG tasks, although not as swiftly as NLU models. Distilled models \cite{ModelCompression} transfer essential knowledge from a larger teacher model to a smaller student model, thus accelerating online inference. Research in information retrieval and contrastive learning has demonstrated the potential for better performance in text representation without the need for labeled data.

In this work, we explore alternatives to the standard pretraining-finetuning paradigm, aiming for a more significant improvement in efficiency while reducing resource costs. Our pretraining scheme, which maintains performance levels, encompasses serving NLU, NLG, and text representation tasks in a zero-shot manner. We propose a simple, efficient, and fine-tune-free framework that is capable of understanding, generating, and representing (GUR) text in a zero-shot manner following unsupervised pretraining.

Our approach enhances the language model by incorporating a contrastive learning task. Given a large general corpus, two sentences extracted from a single document that share a sufficiently long common substring are considered relevant sentences. These sentences form a positive pair for the contrastive learning task, along with in-batch negatives. Subsequently, we pretrain the model using a masked language modeling objective in conjunction with an unsupervised contrastive learning objective.

The resulting model, GUR, demonstrates remarkable performance without the need for any labeled training data points. Notably, it significantly surpasses all other pretrained baselines as a retriever in the recall benchmark under a zero-shot setting and competes closely with BM25 \cite{BM25}, a robust sparse baseline. Moreover, GUR retains its language modeling capabilities, as shown in our ablation experiment. These modules can function separately for different online scenarios, offering increased speed. Once pretrained, GUR is capable of zero-shot learning across various contexts.

%auto-ignore
\section{Related Work}
\label{sec:related}
\subsection{NLP}
Following the release of the Transformer, pretrained models experienced a surge in popularity. GPT\cite{GPT} and BERT\cite{BERT} achieved state-of-the-art (SOTA) results in NLG and NLU tasks. A typical NLP pipeline involves pretraining a large model from scratch or initializing it on a public corpus, followed by post-training on a custom domain corpus, and finally fine-tuning downstream tasks (NLU, NLG, and text representation) on labeled datasets. Some research attempts to simplify this pipeline, with unifying NLP tasks as "text-to-text" yielding notable performance improvements. The primary factor driving these advancements is an increase in model size. Even large models can function as few-shot or zero-shot learners \cite{GPT3}.

Certain studies \cite{2020arXiv200410964G} demonstrate that tailoring a pretrained model to the domain of a target task remains beneficial. \cite{TLM} employs task data as queries to retrieve a small subset of the general corpus and jointly optimizes both the task objective and the language modeling objective from scratch.

Prompt \cite{P3} circumvents the need to fine-tune large models by using prompts to leverage pretrained knowledge. One approach trains a large model \cite{Compress} and subsequently distills it into a smaller model for more efficient inference. Elastic \cite{Elastic} enables early exit for simple samples, thus saving prediction time.

\subsection{Retrieval}
Retrieval is typically perceived as a complex system. BM25 \cite{BM25} serves as a powerful and straightforward sparse retrieval model. Recently, pretraining representation models and interaction models have been incorporated into dense retrieval models \cite{2021arXiv211113853F}. Some research focuses on identifying similar samples, such as the Inverse Cloze Task (ICT) \cite{ICT}, and employs contrastive learning to enhance text representation \cite{Spider}.

\subsection{Representation}
The objective of representation learning \cite{Representation} is to develop an embedding space wherein similar examples are positioned closely together, while dissimilar ones remain distant \cite{Dimensionality}. In contrastive learning, the learning process is formulated as a classification problem, taking into account both similar and dissimilar candidates.

Contrastive learning can be employed in both supervised and unsupervised settings. In the context of unsupervised data, contrastive learning has emerged as one of the most potent approaches in self-supervised learning. \cite{CLIP} utilizes in-batch negative samples, while \cite{ConSERT} and \cite{SimCSE} offer text argument techniques. Some research, such as \cite{seonwoo2023rankingenhanced} and \cite{liu2023rankcse}, employs sophisticated methods to capture more accurate semantic similarity between related sentences.

%auto-ignore
\section{Methodology}
\label{sec:methodology}
\subsection{Pretrain Task}
We incorporate two tasks during pretraining: LM and contrastive learning. Our total loss, denoted by \ref{eqn:loss}, comprises both LM (Language Modeling) loss and CL (Contrastive Learning) loss, as shown below. The variable \textbf{$\alpha$} serves to balance the weight between LM Loss and CL Loss.
\begin{eqnarray}
	\label{eqn:loss}
	Total\,Loss = LM\, Loss + \alpha \, CL\, Loss
	% Total\,Loss = LM\, Loss + α \, CL\, Loss
\end{eqnarray}

% Following BERT\cite{BERT}, we use the MLM (Mask Language Modeling) as our LM object. Besides the LM task, we alse optimize the contrasitive learning object with  the Normalized Temperature-scaled Cross-Entropy (NT-Xent) \cite{SimCLR}. In each batch, we randomly sample $N$ similar text pairs from the corpus while resulting in $2N$ representations as \cite{ConSERT}, \cite{CLIP}. Each sample is trained to find out the similar partner and its counterpart among N(N-1) in-batch negative samples.
% \begin{eqnarray}
% 	% L_{ij} = -log  \frac{  \exp(sim(r_{j},r_{j})/ \tau )  }   { \sum_{k=1}^{2N} I_{k \neq i} \exp( sim(r_{i},r_{k}) / \tau)}
% 	L = -log  \frac{  \exp(sim(r_{i},r_{j})/ \tau )  }   { \sum_{k=1}^{2N} \exp( sim(r_{i},r_{k}) / \tau)}
% \end{eqnarray}
% , where sim(·) indicates the cosine similarity function, τ controls the temperature and I is the indicator. 

In accordance with BERT\cite{BERT}, we employ MLM (Masked Language Modeling) as our LM objective. In addition to the LM task, we also optimize the representation learning objective using a contrastive learning approach. We adopt the same n-pair / InfoNCE \cite{InfoNCE} loss as CLIP, albeit with a fixed temperature of 0.1. This method is slightly different from Normalized Temperature-scaled Cross-Entropy (NT-Xent) \cite{SimCLR} and is adaptive when comparing embeddings from different aspects while consuming less memory.
% L_{ij} = -log  \frac{  \exp(sim(r_{j},r_{j})/ \tau )  }   { \sum_{k=1}^{2N} I_{k \neq i} \exp( sim(r_{i},r_{k}) / \tau)}
% L_{i} = -log  \frac{  e^{ (sim(h_{i},h_{I}) / \tau } )  }   { \sum_{j=1}^{N} e^{ ( sim(h_{i},h_{j}) / \tau) } }
% L = -log  \frac{  \exp(sim(r_{i},r_{j})/ \tau )  }   { \sum_{k=1}^{2N} \exp( sim(r_{i},r_{k}) / \tau)}

Given a batch of $N$ (text, text) pairs, GUR is trained to predict which of the $N \times N$ possible (text, text) pairings across a batch actually occurred. Each sample in a batch has one relevant sample and $N-1$ irrelevant samples. To achieve this, GUR learns a text embedding space by jointly training a shared text encoder to maximize the cosine similarity of the text pair embeddings for the $N$ real pairs in the batch, while minimizing the cosine similarity of the embeddings for the $N*(N-1)$ incorrect pairings. We optimize a symmetric cross-entropy loss over these similarity scores. For a sentence $i$ in a positive relevant pair ($i,I$),  the CL Loss is defined as:
\begin{eqnarray}
	L_{i} = -log  \frac{  \exp{ (sim(h_{i},h_{I}) / \tau } )  }   { \sum_{j=1}^{N} \exp{ ( sim(h_{i},h_{j}) / \tau) } }
\end{eqnarray}
where $\tau$ is the temperature of softmax operation as a hyperparameter.
% where sim(·) indicates the cosine similarity function, τ controls the temperature. 

\cite{2022arXiv220110005N} demonstrates that models often target relevance and semantic textual similarity. For instance, "Tom is chasing Jerry" is relevant to "Jerry is chasing Tom", but their semantics are not equivalent. For a general purpose, we define our contrastive learning objective as relevance. It is preferable to obtain a relevance vector directly and measure semantic textual similarity during fine-tuning tasks.

\begin{figure*}[ht]
	\label{fig:LCS}
	\centering
	\includegraphics[width=0.8\textwidth]{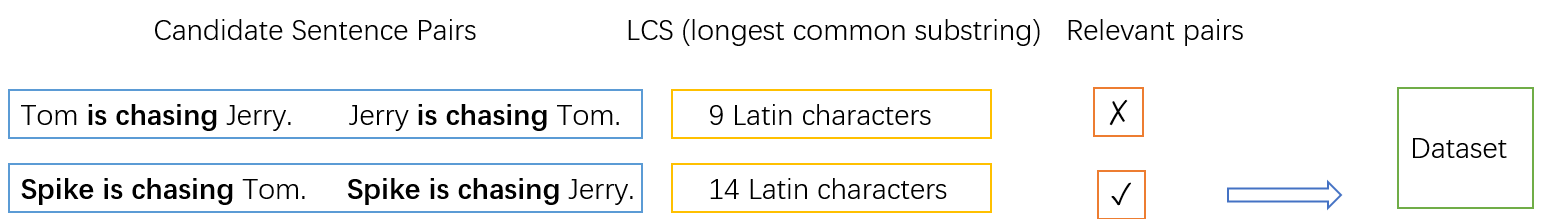}
	\caption{An example of our dataset building approach: For each document, we enumerate sentence pairs as candidates. We then calculate the Longest Common Substring (LCS) for every pair. Pairs with an LCS longer than the threshold are selected as similar texts. We split the article "Tom and Jerry" into sentences and create pairs from every two distinct sentences as candidate similar text pairs. The LCS of "Tom is chasing Jerry." and "Jerry is chasing Tom." is "\textbf{is chasing}", which contains only 9 Latin characters and falls short of our threshold. As a result, they are disregarded as a dissimilar pair. The substring "\textbf{Spike is chasing}" found in the sentences "Spike is chasing Tom." and "Spike is chasing Jerry." serves as the LCS, which consists of 14 Latin characters and is included in our training dataset as a relevant pair.}
\end{figure*}

% \FloatBarrier

\subsection{Corpus}

For pretraining tasks, we exclusively utilize unlabeled data. Few unsupervised methods can compete with BM25. Spider (Span-based unsupervised dense retriever) \cite{Spider} surpasses BM25 by generating numerous query-document pairs from documents. However, its complexity is not ideal for pretraining tasks and results in excessive text fragmentation. Therefore, we simplify this method during pretraining.

As illustrated in \ref{fig:LCS}, the dataset is generated using the following steps:

(1) Divide the document into sentences.

(2) Enumerate all unutilized sentence pairs in a document and generate their normalized versions.

(3) Compute the Longest Common Substring (LCS) of the normalized versions using the \url{https://github.com/laohur/SuffixAutomaton}{SuffixAutomaton}; a pair with a sufficiently long LCS is considered similar.

(4) Incorporate similar pairs into the dataset.

(5) If a sentence's length exceeds the maximum length, it will be randomly cropped \cite{izacard2022unsupervised} in the dataloader.

\begin{figure}[ht]
	\label{fig:Architecture}
	\centering
	\includegraphics[width=0.5\textwidth]{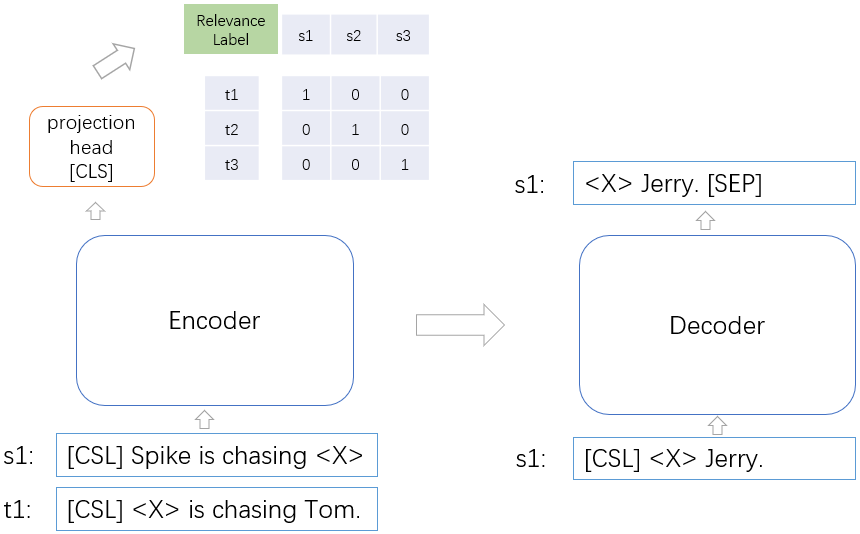}
	\caption{The architecture of our approach. The GUR model, which employs an encoder and a decoder from the Transformer, incorporates a Projection Head to map the sentence embedding to a vector. The related pair $s1$ and $t1$ are flattened in the dataset and subsequently masked as input for T5. The GUR model is designed to jointly train a language model like T5 and predict accurate pairings for a batch of training examples. During testing, the learned text encoder synthesizes a zero-shot classifier by embedding the names of the target dataset's classes.}
\end{figure}

\subsection{Model Architecture}
The scale law suggests that larger models typically exhibit better performance. Our objective is to achieve comparable performance while enhancing training and inference speed. The framework should facilitate Natural Language Understanding (NLU), Natural Language Generation (NLG), and text representation tasks with a zero-overhead abstraction mechanism. Consequently, our framework comprises a Transformer encoder, a Transformer decoder, and a projection head for representing sentences as vectors. These components can be employed independently for various tasks during both training and inference stages.

Our model, as depicted in \ref{fig:Architecture}, is based on T5 for a shorter decoding length. To enhance inference speed, we refrain from formatting all tasks into text-to-text, as in the original T5 style. Instead, we utilize only the T5 Encoder for processing NLU tasks. Some studies (\cite{CPT}, EncT5\cite{EncT5}) suggest that the encoder plays a more crucial role in NLU tasks than the decoder. Thus, employing just the encoder in a Transformer model is both sufficient and expedient.

A pre-trained model is adaptable for various downstream tasks, thanks to extensive training datasets. We introduce a projection head above the base encoder to represent sentences as fixed-dimensional vectors. As \cite{SimCLR} suggests, the contrastive task is trained using a non-linear projection head. We incorporate a non-linear projection head to represent a sentence as a fixed-dimensional vector. To expedite text representation and avoid conflicts between multiple objectives, unlike BART, the projection head is positioned above the encoder rather than the decoder. This configuration allows the tensor to flow solely through the encoder during inference.

%auto-ignore
\section{Experiments}
\label{sec:experiment}
\subsection{pre-training}
\label{ssec:pre-training}
We first pretrain a small-sized model, as demonstrated in \ref{tab:Setting}. GUR-Small comprises 8 encoder layers and 8 decoder layers and is initialized using the small-sized model available at \url{https://huggingface.co/IDEA-CCNL/Randeng-T5-Char-57M-MultiTask-Chinese}{Hugging Face}. We incorporate a projection head after the encoder to encode sentence representations. The projection head maps the "[CLS]" token representation from the final encoder layer output to a 128-dimensional vector. The weight of the Contrastive Learning (CL) Loss, $\alpha$, is set to 1, ensuring that the CL Loss and Masked Language Model (MLM) Loss values remain reasonably close.

\begin{table}[h]
      \centering
      \caption{ Pre-training Setting}
      \label{tab:Setting}
      \begin{tabular}{cc}
                             & GUR-Small            \\ \hline
            optimizer        & AdamW                \\
            lr scheduler     & constant with warmup \\
            learning rate    & 1e-4                 \\
            masking rate     & 15\%                 \\
            $\alpha$         & 1                    \\
            temperature      & 0.1                  \\
            model dim        & 512                  \\
            vector dim       & 128                  \\
            encoder layers   & 8                    \\
            decoder layers   & 8                    \\
            projection token & {[}CLS{]}            \\
            tokenizer        & WordPiece            \\
            % vocab            & 12902
      \end{tabular}
\end{table}

Owing to resource constraints, we choose to sample a masked sentence span for the model input rather than employing more complex text augmentation methods. We utilize the Pairwise Ranking Objective for model representation instead of the Listwise Ranking Objective.

\begin{figure}[h]
      \centering
      \includegraphics[width=\linewidth]{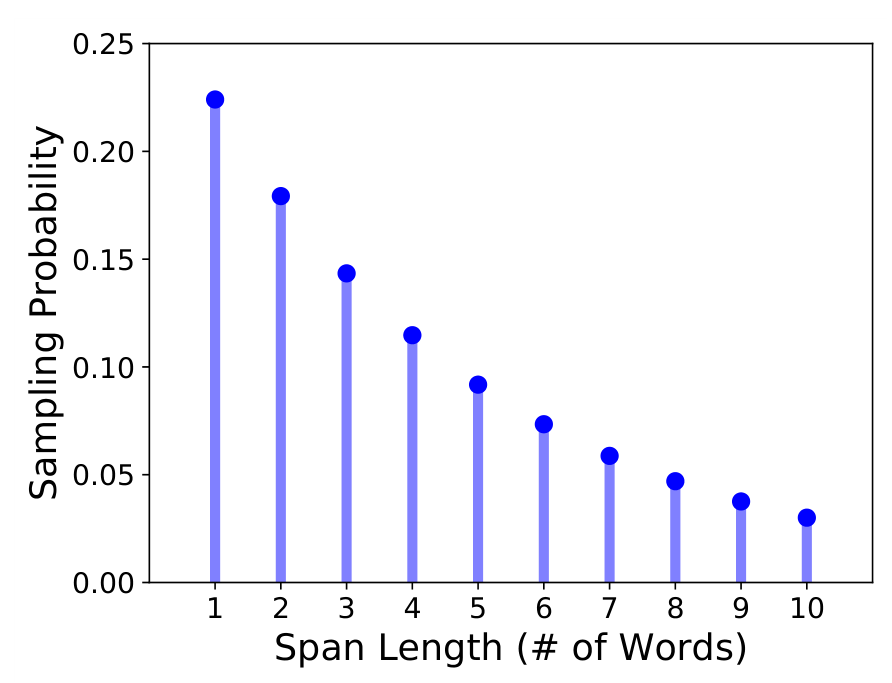}
      \caption{ SpanBERT\cite{SpanBERT} samples random span lengths from a geometric distribution $L\sim$ Geo($p$ = 0.2) clipped at $L_{max} = 10$.    }
\end{figure}

The masking rate is set at 15\%. SpanBERT\cite{SpanBERT} employs a geometric distribution for its masking distribution, which is skewed towards shorter spans. However, many of our sentences contain no more than ten words. The default masking strategy of SpanBERT generates numerous one-token mask spans, causing the masking rate to potentially deviate from the target masking rate. To address this issue, we sample our masking distribution from a geometric progression with a peak, tailored for our fragmented sentences. This approach proves to be more robust for short sentences.
\begin{eqnarray}
      \label{eqn:hump}
      P(k;p;mode) = Normalize(p^{abs(k-mode)})
\end{eqnarray}

In the \textbf{Hump Geometric Distribution}, the probability decreases as the distance from the peak increases, as illustrated in \ref{eqn:hump}. This strategy favors generating longer spans compared to the geometric distribution, while maintaining the same expected value.
\begin{figure}[h]
      \centering
      \includegraphics[width=\linewidth]{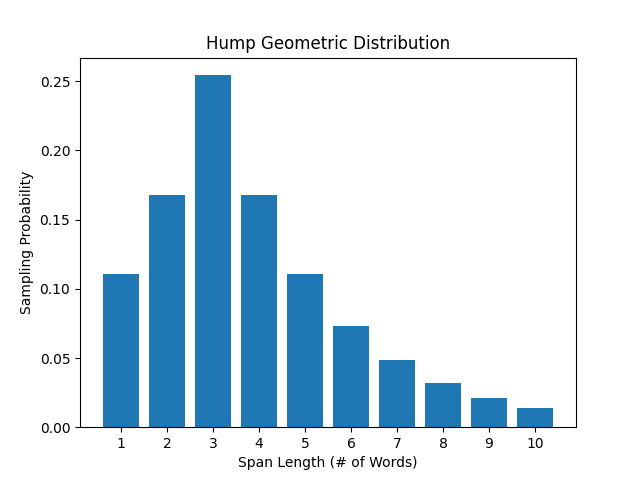}
      \caption{We sample random span lengths from a hump geometric distribution $L\sim$ HumpGeo($p$ = 0.66, $mode$=3) and clip them at $L_{max} = 10$. This approach is more robust for short sentences.}
\end{figure}

\begin{lstlisting}[   % 进行参数设置
      language=Python, % 设置语言
      breaklines=true, % 自动换行
      float
     ]  
# Hump Geometric Distribution
distribution = [p ** abs(i - mode) for i in range(lower, upper + 1)]
distribution = [x / (sum(distribution)) for x in distribution]
# keep lower=1, upper=10, mean=3.8 as SpanBERT
# set p=0.66 as a hyper parameter, mode=3 as T5
\end{lstlisting}

Our corpus comprises Wikipedia \url{https://dumps.wikimedia.org/}{dumps} acquired using \url{https://github.com/laohur/wiki2txt}{wiki2txt}, CSL \cite{li-etal-2022-csl}, and custom documents. Notably, our texts are often of low quality, with some documents containing only a single high-quality sentence in the title amidst the document content. For each document, we sequentially enumerate every content sentence in relation to the title, forming potential similar text pairs. Ultimately, the large dataset is deduplicated and shuffled using bigsort, accessible at \url{https://github.com/laohur/bigsort}{GitHub}, and read in a stream to minimize resource requirements.

To expedite the pre-training phase and augment the number of training steps, we divide the dataset based on sentence length. Sentences with a length of \textless{} 64 (lcs \textgreater{}= 2 Hanzi) are trained using a sequence length of 32. Sentences with a length of \textgreater{}= 64 (lcs \textgreater{}= 3 Hanzi) are trained using a sequence length of 128. The batch size varies according to the sequence length. Additionally, we incorporate a prompt task called "document2title" for the generation task.

For the ablation study, we pre-trained four models under different conditions. The "GUR-FULL" or "GUR" model integrates all the methods mentioned earlier, serving as our foundational model. The "GUR-LCS" model processes candidate pairs without LCS filtering, which is akin to ICT but less random in this experiment. The "GUR-LM" model omits the LM task, while the "GUR-CL" model removes the CL task.

%auto-ignore
\section{Result}
\label{sec:result}

\subsection{retrieval}
\label{ssec:retrieval}

\begin{figure}[h]
      \label{fig:recall100}
      \centering
      \includegraphics[width=0.5\textwidth]{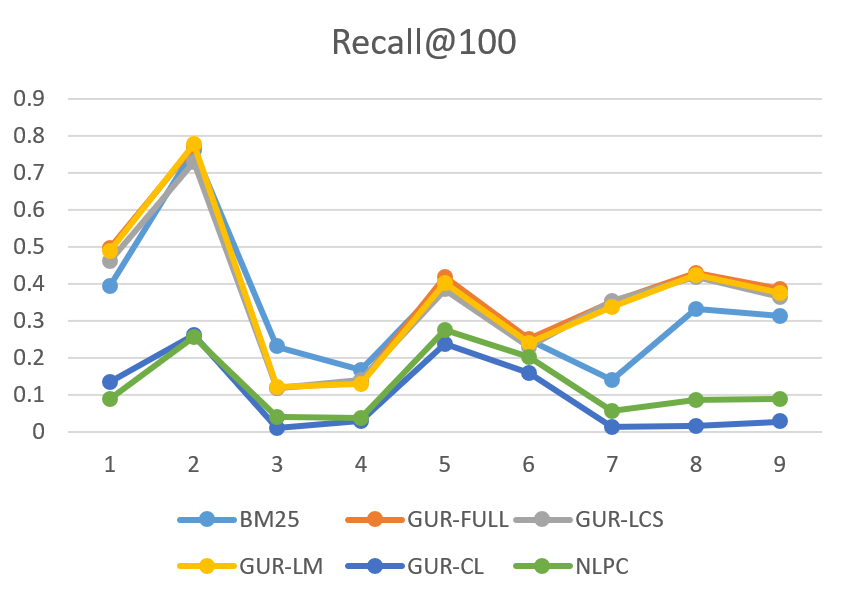}
      \caption{The recall@100 performance of the models on retrieval tasks. The task score curves among the models are nearly parallel. Only BM25 stands out in task3 and underperforms in task7, task8, and task9, due to term hits or mismatches.}
\end{figure}

\begin{figure}[h]
      \label{fig:MRR10}
      \centering
      \includegraphics[width=0.5\textwidth]{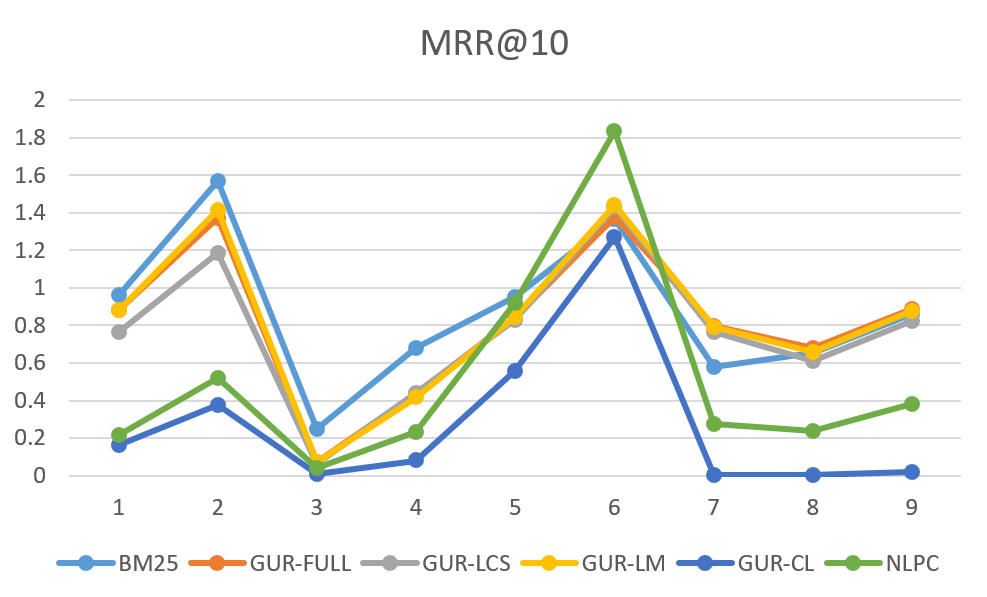}
      \caption{The MRR@10 performance of the models across retrieval tasks. Some tasks are either extremely challenging or easy for all models.}
\end{figure}

We employ recall@100 and MRR@10 scores in retrieval tasks to evaluate the text representation results. In addition to the "GUR" models, BM25 and NLPC (\url{http://nlpc.baidu.com/platform/demo/wordemb}) also participate in these retrieval tasks without any fine-tuning. We simply obtain sentence (dense or sparse) embeddings from the model outputs and use similar candidates as retrieval results. Although not mandatory, we extract the GUR Encoder and Projector to create GurForSequenceRepresentation, which embeds the text of "GUR" models for reduced GPU memory consumption.

Table \ref{fig:recall100} displays the recall performance across different tasks. The "GUR-FULL," "GUR-LM," and "GUR-LCS" models perform almost identically and outperform the others. In this experiment, our LM and CL tasks exhibit no conflict. Since we generate similar pairs by coupling document titles with sentences in the document content, the datasets used in the "GUR-FULL" and "GUR-LCS" models exhibit slight differences. The "BM25" model excels in task3 but lags in task7, task8, and task9 due to term hits or mismatches. The "GUR-CL" and "NLPC" models obtain the lowest scores because the "GUR-CL" model is trained without the CL task, and the "NLPC" model relies on Word2Vec \cite{2013arXiv1301.3781M}, \cite{2013arXiv1310.4546M}.

We also employ MRR@10 \ref{fig:MRR10} to more accurately measure text representation. The MRR@10 score curves for these models exhibit minor differences compared to the recall@100 score curves. The difficulty between recall and MRR varies across some tasks. Certain tasks are challenging to recall but easier to rank. The sparse model "BM25" performs best in most tasks due to precise term hits. The "GUR-FULL" and "GUR-CL" models closely follow. The "GUR-LCS" model experiences a slight decline in some tasks. The simpler "NLPC" model achieves the highest score in a few tasks.

In the retrieval benchmark, we observed that the LM task and CL task can be trained concurrently, even for a small-sized model. No single model excels in both recall (GUR) and ranking (BM25) performance.

\subsection{NLU}
\label{ssec:NLU}

\begin{figure}[h]
      \label{fig:Zero-shot}
      \centering
      \includegraphics[width=0.5\textwidth]{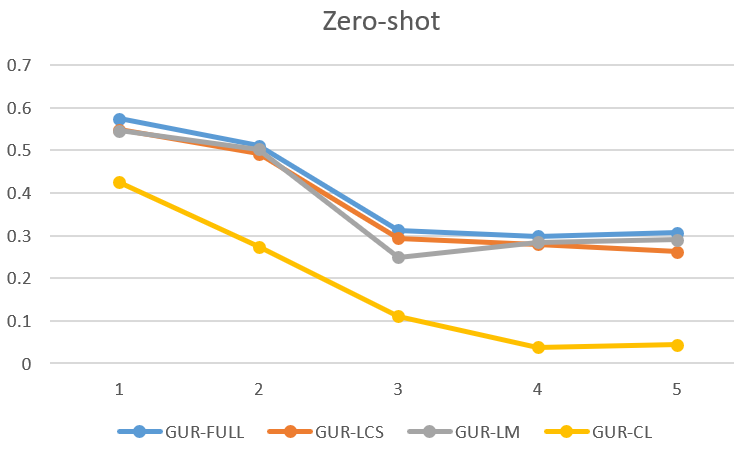}
      \caption{Multi-classification ACC results on zero-shot tasks. Each task comprises tens or hundreds of classes. Most labels consist of one or two terms. The models encode both the samples and the labels into vectors, subsequently searching for the nearest label for each sample.}
\end{figure}

\begin{figure}[h]
      \label{fig:few-shot}
      \centering
      \includegraphics[width=0.5\textwidth]{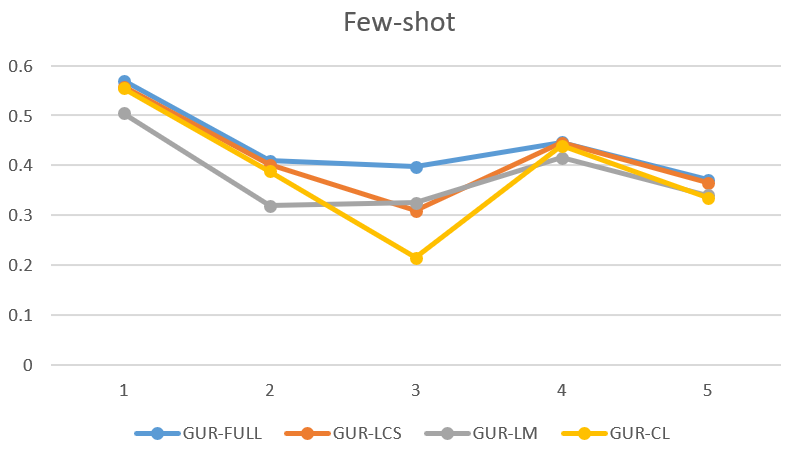}
      \caption{Multi-classification ACC results on few-shot tasks. Each task comprises tens or hundreds of classes. In the training dataset, each class contains 10 samples. All models are trained under the same conditions. The varying scores indicate the models' ability to handle a limited number of samples.}
\end{figure}

\begin{figure}[h]
      \label{fig:Fine-tune}
      \centering
      \includegraphics[width=0.5\textwidth]{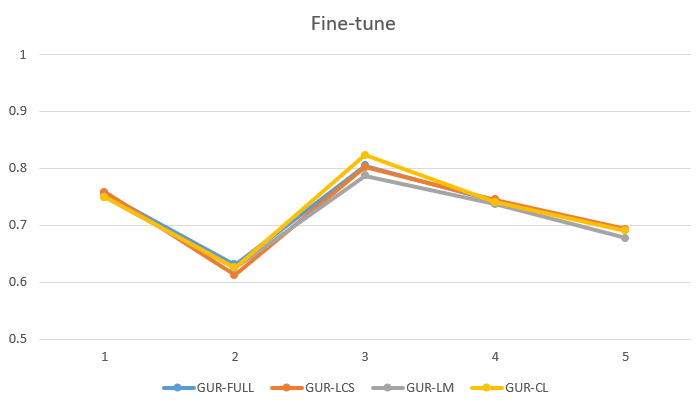}
      \caption{Multi-classification ACC results on fine-tuning tasks. Each task comprises tens or hundreds of classes. In the training dataset, each class contains 100 samples. All models are trained under the same conditions. Nearly all models achieve the same score after sufficient fine-tuning. The scores across the tasks indicate the difficulty level of each task.}
\end{figure}

We assemble several custom text classification tasks to evaluate the models in zero-shot, few-shot, and fine-tuning scenarios. Each task comprises tens or hundreds of classes.

In each zero-shot task, every sample is associated with a meaningful text label. The models transform queries and labels into vectors and identify the nearest label among hundreds of options for each query in a zero-shot manner. We simply extract the GUR Encoder to create GurForSequenceClassification, akin to BertForSequenceClassification, for classifying queries during the fine-tuning process. As depicted in \ref{fig:Zero-shot}, the "GUR-FULL" and "GUR-LM" models achieve the best results, with the "GUR-LCS" model closely following. Naturally, the "GUR-CL" model, which lacks CL task pre-training, falls significantly behind. These results demonstrate that the LM and CL tasks can simultaneously contribute to the pre-trained model.

The majority of models achieve comparable scores after undergoing sufficient training in identical environments, as illustrated in Figure \ref{fig:Fine-tune}. These scores reflect the inherent complexity of the tasks. As the task difficulty varies, so does the gap between fine-tuning and zero-shot performance. Considering a limited number of samples in the training dataset, Figure \ref{fig:few-shot} demonstrates the effectiveness of models in few-shot scenarios. Among all tasks, the "GUR-FULL" model emerges as the most outstanding performer.

\subsection{NLG}
\label{ssec:NLG}

\begin{table*}[ht]
      \caption{Natural Language Generation Results for NLG Tasks: Translated from Chinese to English. The models generate the masked spans within the prompts, replacing "\textless{}{[}MASK{]}\textgreater{}" with the generated span. Due to poor quality and excessive length, the "GUR-LM" results are disregarded. Models pretrained with the LM task exhibit similar and fluent generated outcomes. By masking random optional words, we transform Artificial Prompts into Auto Prompts. The resulting sentences from Auto Prompts display greater diversity compared to those derived from the Artificial Prompts.}
      \label{tab:Prompts}
      \begin{tabular}{lllll}
            \textbf{Auto Prompts}                                                                      & \textbf{Artificial Prompts}                                                                                \\
            \multicolumn{2}{c}{\textit{GUR-FULL}}                                                                                                                                                                   \\
            Chicken Manure \textless{}Organic   Fertilizer\textgreater Separator Principle             & What brand of \textless{}solid-liquid\textgreater separator is of good quality                     &  &  & \\
            Ground flat iron anticorrosion   \textless{}paint\textgreater{}                            & What is flat iron \textless{}angle steel\textgreater{}                                             &  &  & \\
            Standard height of welding platform   \textless{}how much\textgreater{}                    & \textless{}Resistance\textgreater What is the welding platform?                                    &  &  & \\
            Dump truck transportation of stone   \textless{}what is\textgreater{}?                     & Specific introduction about dump trucks                                                            &  &  & \\
            How much can you drive a long auger in   \textless{}Line Plus\textgreater{}                & Which \textless{}straight\textgreater auger is easy to use                                         &  &  & \\
            What is composite PVC sports floor   \textless{}material\textgreater{}                     & \textless{}Outdoor\textgreater What is the sports floor                                            &  &  & \\
            The front room of the elevator is   \textless{}whether there\textgreater air supply outlet & What is the vent made of                                                                           &  &  & \\
            What does the top support of seawater   \textless{}pt is\textgreater mean?                 & Where is jacking \textless{}mainly\textgreater used                                                &  &  & \\
            Vehicle-mounted aerial work platform                                                       & Introduce \textless{}what is\textgreater aerial work platform                                      &  &  & \\
            \textless{}Greenhouse\textgreater Can black mulch be   used?
                                                                                                       & \textless{}Black\textgreater black mulch \textless{}choice\textgreater what brand is good          &  &  & \\
            \multicolumn{2}{c}{\textit{GUR-LCS}}                                                                                                                                                                    \\
            Principle of chicken manure   \textless{}solid-liquid\textgreater separator                & What brand of \textless{}oil-water\textgreater separator is of good quality                        &  &  & \\
            Ground flat iron anti-corrosion   \textless{}wood\textgreater{}                            & What is the flat iron \textless{}model number\textgreater{}                                        &  &  & \\
            Standard height of welding platform   \textless{}is more\textgreater{}                     & \textless{}Rail\textgreater What is the welding platform?                                          &  &  & \\
            How to transport stone by dump truck   \textless{}method\textgreater{}?                    & Specific introduction about dump trucks                                                            &  &  & \\
            How much can a long auger run in   \textless{}under water\textgreater{}                    & Which \textless{}spiral\textgreater auger is easy to use                                           &  &  & \\
            What is composite PVC sports floor   \textless{}material\textgreater{}                     & \textless{}Outdoor\textgreater What is the sports floor                                            &  &  & \\
            The front room of the elevator is   \textless{}whether there\textgreater air supply outlet & What is the vent made of                                                                           &  &  & \\
            What does the top support of sea water   \textless{}yes\textgreater mean?                  & Where is jacking \textless{}mainly\textgreater used                                                &  &  & \\
            Vehicle-mounted aerial work platform                                                       & Introduce \textless{}outdoor\textgreater aerial work platform                                      &  &  & \\
            \textless{}Vegetable field\textgreater Can black mulch   be used?
                                                                                                       & \textless{}PE\textgreater{}black mulch \textless{}machine selection\textgreater what brand is good &  &  & \\

            \multicolumn{2}{c}{\textit{GUR-CL}}                                                                                                                                                                     \\
            Principle of chicken manure \textless{}dry and   wet\textgreater separator                 & What brand of \textless{}solid-liquid\textgreater separator is of good quality                     &  &  & \\
            Ground flat iron anticorrosion   \textless{}treatment\textgreater{}                        & What is a flat iron \textless{}cross bar\textgreater{}                                             &  &  & \\
            Standard height of welding platform   \textless{}how much\textgreater{}                    & \textless{}Car\textgreater What is the welding platform?                                           &  &  & \\
            Self-dumping car   to transport stone \textless{}does it\textgreater{}?                    & Specific introduction about dump trucks                                                            &  &  & \\
            How much can a long auger drill run in   \textless{}Jiangxi\textgreater{}                  & Which \textless{}crawler\textgreater auger is easy to use                                          &  &  & \\
            What is composite PVC sports floor   \textless{}Material\textgreater{}                     & \textless{}Rubber\textgreater What does sports flooring do?                                        &  &  & \\
            The \textless{}what\textgreater air supply vent is in   the front room of the elevator     & What is the vent made of                                                                           &  &  & \\
            What does the top support of sea water   \textless{}yes\textgreater mean?                  & Where is jacking \textless{}mainly\textgreater used                                                &  &  & \\
            Vehicle-mounted aerial work platform                                                       & Introduce \textless{}those\textgreater aerial work platforms                                       &  &  & \\
            \textless{}Potatoes\textgreater Can black mulch be   used?                                 & \textless{}wheat\textgreater black mulch \textless{}material\textgreater which brand is good       &  &  &
      \end{tabular}
\end{table*}

% The prompts are generated by randomly mask non-keywords in querys in left column. The results generated by models in right column are used as query expansion. 
We assess the Natural Language Generation (NLG) capabilities of models in a zero-shot query expansion task using two distinct approaches for generating prompts. Initially, custom keywords are employed to retrieve queries, which are then masked, excluding the keywords, to create auto prompts. In contrast, artificial prompts are composed by human writers. These prompts are fed into the models, and the generated outcomes are subsequently analyzed and compared.

Table \ref{tab:Prompts} displays samples of both auto prompts (left column) and artificial prompts (right column). Upon reviewing the results in an artificial setting, it appears that models pretrained with the LM task exhibit similar performance. The majority of samples are suitable for query expansion, and results derived from artificial prompts demonstrate greater controllability. Interestingly, auto prompts, generated through random masking, perform comparably to artificial prompts while showcasing increased diversity.

%auto-ignore
\section{Conclusion}
\label{sec:conclusion}

This study presents an alternative approach to the traditional pretraining-finetuning paradigm, with the goal of significantly enhancing efficiency at a reduced cost. We introduce a straightforward, effective, and finetune-free framework that can understand, generate, and represent (GUR) text in a zero-shot manner following unsupervised pretraining. Our pretraining method relies on instances of similar text pairs selected based on their longest common substring (LCS) from unannotated documents. This approach combines masked language modeling with an unsupervised contrastive learning task. Experimental results demonstrate that GUR achieves comparable performance to pre-trained language models (PLMs) in natural language understanding (NLU) and natural language generation (NLG) tasks while outperforming BM25 in recall tasks as a dense retriever in a zero-shot setting.  The model architecture has been tailored for inference optimization across diverse scenarios by leveraging individual modules.

%auto-ignore
\section{Limitation}
\label{sec:limitation}
Due to resource constraints, our study has only been able to conduct limited experiments under specific conditions using a custom corpus. We have tested the framework with verified settings and restricted resources. A more comprehensive investigation would involve pretraining from scratch, benchmarking on general tasks, large-scale evaluations, and exploring more accurate and efficient sentence representations and contrastive approaches. Some of these investigations are currently underway. Our aim is to contribute to a more democratic pretraining style for AI models through this work.

\FloatBarrier
\bibliographystyle{IEEEtran}
% \bibliography{bibliography/anthology,bibliography/eacl2021,bibliography/nlp,bibliography/retrieval,bibliography/represent}
\bibliography{bibliography/nlp,bibliography/retrieval,bibliography/represent}

\end{document}